%% file: main.tex
\documentclass[letterpaper]{article} %DO NOT CHANGE THIS
\usepackage{aaai18}  %Required
\usepackage{times}
\usepackage{helvet}
\usepackage{courier}
\usepackage{graphicx}
\usepackage{amsmath}
\usepackage{amsthm}
\usepackage{amssymb}
\usepackage{caption,subcaption}
\usepackage{wrapfig}
\usepackage{titlesec}
\usepackage{hyperref}
\usepackage{mathabx}
\usepackage{amsthm}

\theoremstyle{definition}
\newtheorem{definition}{Definition}

\usepackage{caption,subcaption}

\begin{document}
\title{Domain Aware Markov Logic Networks}
\author{Happy Mittal, Ayush Bhardwaj \\ 
	Dept. of Comp. Sci. \& Engg. \\
	I.I.T. Delhi, Hauz Khas \\ 
	New Delhi, 110016, India \\
	\texttt{\scalebox{0.9}{happy.mittal$@$cse.iitd.ac.in}},\\
	\texttt{\scalebox{0.9}{ayushb647$@$gmail.com}} \\
	\And Vibhav Gogate \\ 
	Dept. of Comp. Sci. \\ 
	Univ. of Texas Dallas \\
	Richardson, TX 75080, USA \\
	\texttt{\scalebox{0.9}{vgogate$@$hlt.utdallas.edu}} \\ 
	\And Parag Singla \\ 
	Dept. of Comp. Sci. \& Engg. \\
	I.I.T. Delhi, Hauz Khas \\
	New Delhi, 110016, India \\
	\texttt{\scalebox{0.9}{parags$@$cse.iitd.ac.in}} \\
}
\maketitle
\input{abstract}

\input{intro}
%\input{background}
%\input{problem_mln}
%\input{problem_mln_2}
%\input{problem_mln_3}
\input{agg}
\input{experiments}
\input{ack}
%\input{conclusion}
%\input{Theoritical_limits}
%\input{Theoritical_limits_1}
%\input{Theoritical_limits_2}
\bibliographystyle{aaai}
%\newpage
\bibliography{all}

\end{document}

%% file: abstract.tex
Combining logic and probability has been a long standing goal of AI research. Markov Logic Networks (MLNs) achieve this by attaching weights to formulas in first-order logic, and can be seen as templates for constructing features for ground Markov networks. Most techniques for learning weights of MLNs are domain-size agnostic, i.e., the size of the domain is not explicitly taken into account while learning the
parameters of the model. This often results in extreme probabilities when testing on domain sizes different from those seen during training. In this paper, we propose {\it Domain Aware Markov logic Networks} (DA-MLNs) which present a principled solution to this problem. While defining the ground network distribution, DA-MLNs divide the ground feature weight by a scaling factor which is a function of the number of connections the ground atoms appearing in the feature are involved in. %each ground atom in the feature is involved in.
%by dividing the ground feature weight by a function of the number of connections each ground atom (in the feature) is involved in, when
%defining the ground Markov network distribution. 
We show that standard MLNs fall out as a special case of our formalism when this function evaluates to a constant equal to 1.
%is a constant (and is equal to 1).  
Experiments on the benchmark Friends \& Smokers domain show that our approach results in significantly higher accuracies compared to existing methods when testing on domains whose sizes different from those seen during training.

%% file: intro.tex
\section{Introduction}
%Several real world problems in AI have relational structure among entities as well as uncertainty in the relations which can be aptly represented by Statistical relational learning (SRL) models. Several SRL models~\cite{getoor&taskar07} have been proposed in the literature, which combine logical representations and statistical models to represent the underlying data. 
Markov Logic~\cite{domingos&lowd09} is a powerful Statistical Relational Learning (SRL) formalism, which represents the underlying domain using weighted first-order logic formulas. Markov Logic has been successfully applied to a number of problems including those in information extraction, NLP, social network analysis, robot mapping and computational biology~\cite{domingos&lowd09}. Most existing applications of Markov logic implicitly assume that test data is similar in size to the training data, and hence, weights learned on the training data can naturally be used for prediction on the test data. But for many real world settings, this assumption may not hold true any more, simply because the high costs of annotation may require us to learn the model on a relatively small sized training data, while still requiring prediction on large test sizes.
%To our best of knowledge, in all the applications where MLNs have been used, size of training and test data were almost similar, and so weights learned from training data were naturally adapted well for test data. However, in most of the real problems, we don't have much annotated data due to which available training data is very small as compared to test data. In those cases, weights learned from training data may not be suitable for test data of very large size. 

To illustrate the problem, suppose we have an MLN with a single formula $w : P(x,y)\Rightarrow Q(x)$, where $w$ is the weight learned using some training data. If domain size of $y$ increases in the test data, then for a particular grounding $q$ of predicate $Q$, number of its neighbors (i.e., groundings of $P$), also increase, and it can be shown that the combined effect of all the neighbors  results in extreme marginal probability being assigned to $q$. Poole et al. ~\shortcite{poole&al14} characterized this behavior for some classes of MLNs, but they did not provide any solution to the problem.
%  models represent both relational structure and uncertainty in the underlying domain.
% In most of the real problems, size of training data is very less as compared to size of test data. This is primarily due to lack of labeled data. 
So the question is how to \emph{transfer} the weights learned from training data of a certain size to the weights suitable for test data of a different size. Jain et al.~\shortcite{jain&al10} proposed Adaptive Markov
Logic Networks (AMLNs) in which weights are learned over multiple databases of different sizes, and these weights are then approximated using a linear combination of pre-defined set of basis functions. Their approach has mainly two limitations : (a) The basis functions (hence the weights) depend only on the size of the domain, whereas as described above, we would like to focus on the number of neighbors of ground atoms in a formula 
%(rather than the total groundings of the formula), 
%(b) There is a single weight transformation learned for entire MLN, which means each weight in MLN is transformed in a similar manner, but in general we may not want to do so. 
(b) They do not provide any strong (theoretical) justification for why their basis function approach should work.
%in practice. 
%ir basis functions do not to have any justification as to why they should work. 
Further, in our experiments, we did not find their approach suited to benchmark problems even when compared to standard MLNs.
% even when compared to standard MLNs.
%problems. 
%MLNs.
%even when compared to standard MLNs. 
%their approach performed poorly on the benchmark domains we experimented on.
%some of the benchmark MLN domains.
%Experimentally, we found their basis functions to be not very suitable for real world problems.
%too weak for real world datasets.
%We found For example, suppose we add another formula in our example MLN : $R(x)\Rightarrow Q(x)$, then a particular ground atom $q$ of $Q$ will be affected only from one ground atom of $R$, so for this formula, we don't need to transform the weight for different domain size. 

In this work, we propose a modified MLN formalism, called {\em Domain Aware Markov Logic Networks (DA-MLNs)}, in which weights are dynamically adapted to different domain sizes based on some of the ideas described above.
%according to the ideas we described above. 
%In the following section, we describe DA-MLNs, and then we experimentally show on a benchmark dataset that DA-MLNs outperform traditional MLNs and Adaptive MLNs.
%They are called adaptive because the distribution defined over the Markov Network by them adapts itself according to size of the underlying data. 
%As the name suggests, the new distribution would adapt itself according to the size of underlying data. The key intuition which we have used in OMLNs is that traditional MLNs do not incorporate size of the data into weights of the formula. So in AMLNs, we augment the weights with this extra information which enables it to adapt the distribution according to the data size.

%% file: agg.tex
\section{Domain Aware Markov Logic Networks (DA-MLNs)}
\label{section:agg}
As described in the previous section, in Markov Logic, marginal probability of a grounding $q$ of a first-order predicate $Q$ tends to extreme as the cumulative effect of number of ground formulas in which $q$ appears increases. 
%Here, we describe an approach to explicitly incorporate the effect of number of connections of $q$ in the distribution. 
A natural solution to this problem is to \emph{scale down} the effect of each ground formula (in which $q$ appears), so that even in very large domains, cumulative effect would not result in extreme probabilities. 
%almost remain constant. 
Intuitively, this scaling factor should depend on the number of connections the ground atoms in a formula are involved in.
%(and hence the corresponding predicate). 
To formalize this notion, 
%scaling factor, 
we first define number of connections of a first-order predicate.
%define \textbf{aggregation factor} of a ground formula to capture the effect of domain size on the MLN weights. In a MLN, as each grounding is a separate feature, we can afford to have different weights for each grounding of a first order formula. First, we will introduce  num\_connections for a ground predicate, followed by num\_connections for a grounding of First Order Formula. Finally we would define the aggregation factor for a particular grounding. Let $F_{i}$'s be the First Order Formulas.
%\begin{definition}\textbf{(Num\_Connection of ground atom)} Let $F$ be a First order formula. Let $P$ be a ground atom appearing in some grounding of $F$. Then Number of Connections of $P$, denoted by $NC(P)$, is defined as the number of times $P$ occurs in the groundings of $F$ at the same position.
%\end{definition} 
%\begin{definition}\textbf{(Num\_Connection of ground formula)} Let $F$ be a First order formula. Let $G_{j}$ be some ground formula of $F$. Let $P_{1}, P_{2}, ... P_{m}$ be the ground atoms present in $G_{j}$. Then Number of Connections of $G_{j}$, denoted by $NC(G_{j})$ is defined as a vector $\{NC(P_1),NC(P_2),\ldots,NC(P_m)\}$.
%%\\ Num\_Connection($GF_{i,j}$) = Num\_Connection($P_{1}$), Num\_Connection($P_{2}$), ... Num\_Connection($P_{m}$)
%\end{definition}
\begin{definition}
	\textbf{(NumConnections)}  Let $F$ be a first order formula containing predicates $[P_1,P_2,\ldots P_m]$. Let $Vars(P_j)$ be the set of logical variables appearing as arguments of  $P_j$. Let $Vars(P_j)^-$ denote the set of all the logical variables in $F$ not appearing in $P_j$. Then, we define NumConnections of $P_j$ , denoted by $c_j$ as, $\max\left(1, \prod_{x\in Vars(P_j)^-} |\Delta x|\right)$, 
	where $\Delta x$ is the domain of variable $x$. Intuitively,  $c_j$ is the number of ground formulas which affect marginal probability of any instantiation of $P_j$.
\end{definition} 
Since each predicate in a formula can have different number of connections, we define a connection-vector of a first order formula as : 
\begin{definition}
	\textbf{(Connection-vector)} Let $F$ be a first-order formula containing predicates $[P_1,P_2,\ldots, P_m]$. Then connection-vector $v$ is defined as $[c_1,c_2,\ldots,c_m]$ where each $c_j (1 \le j \le m)$ denotes $NumConnections$ for $P_j$. 
\end{definition}
\textbf{Example : } Consider again our example formula $w : P(x,y) \Rightarrow Q(x), \Delta x = \Delta y = \{a,b\}$. Then its connection vector $v$ would be $[1,2]$. Therefore, the connection-vector of a formula captures the number of connections for each of its predicates. Higher the number of connections in a formula, higher should be the scaling factor for that formula. %So our scaling factor should be directly proportional to the number of connections. 
We define our scaling factor as : 
\begin{definition}\textbf{(Scaling-down Factor)} Let $F$ be a first-order logic formula. Let $v$ be its connection vector. Then its scaling-down factor, $s = \max(v)$, where $\max(v)$ returns maximum element of vector $v$.
\end{definition}
Though we have chosen $\max$ function to capture the effect of number of connections in the definition above, other functions such as $\sum$ could also be used. Empirically, we found $\max$ to perform well. 
%Exploring other functions in the spirit of ~\cite{} is a direction for future work. 

Next, we describe the probability distribution defined by the DA-MLN model. Given an assignment $x$ to all the ground atoms $\mathbf{X}$,
%$\mathbf{X}$ in a DA-MLN $M$, the 
the probability of $\mathbf X=x$ is given by: 
$$P(\mathbf{X}=x;w) = \frac{1}{Z}\exp\left(\sum_{i=1}^{n} \frac{w_{i}}{s_i} n_{i}(x)\right)$$
where $w_i$, $s_i$ and $n_i$ denote the weight, scaling factor, and the number of true groundings of $i^{th}$ formula, respectively. 
DA-MLNs differ in the way their probability distribution is defined. They subsume (become identical to) MLNs if we replace $s_i$ by 1 for all the formulas. In our analysis, we have been able formally prove that at least for some simple class of formulas, DA-MLNs do not result in extreme marginal probabilities. We plan to publish these results in an extended conference version of the paper. Exploring connections with work on aggregators in relational models~\cite{kazemi&al17} is a direction for future work.

%% file: experiments.tex
\section{Experiments}
We implemented DA-MLNs on top of the existing Alchemy~\cite{kok&al07} system. For AMLNs~\cite{jain&al10}, the weights learned using Alchemy were used to learn the coefficients of the basis functions.
For MLNs, we used the default Alchemy implementation. For each of the models, CG was used as the learning algorithm, and Gibbs sampling was used for inference. In each case, default parameter settings were used.

We compared each of the three approaches on the standard Friends \& Smokers (FS) domain~\cite{singla&domingos08}. The domain has two rules: smoking leads to cancer and friends have similar smoking habits (along with singleton for each predicate). Since we would like to predict on varying domain sizes, we generate the data with domain size of $n$ as follows: we randomly create $\sqrt{n}$ number of (equal-sized) groups, such that people in each group are more likely to be friends with each other ($p_f=0.8$), whereas people across groups are less likely to be friends with each other ($p_f=0.1$). Each group is randomly decided to be a smoking group with probability $p_g=0.3$. In a smoking group, each person smokes with probability of $p_s=0.7$, and in a non-smoking group, each person smokes with probability $p_s=0.1$. A smoking person has cancer with probability $p_c=0.5$, and a non-smoking person has cancer with probability $p_c=0.01$.

\begin{figure}[h]
 	\centering
 	\begin{subfigure}[]{0.35\textwidth}
 		\includegraphics[width=\linewidth]{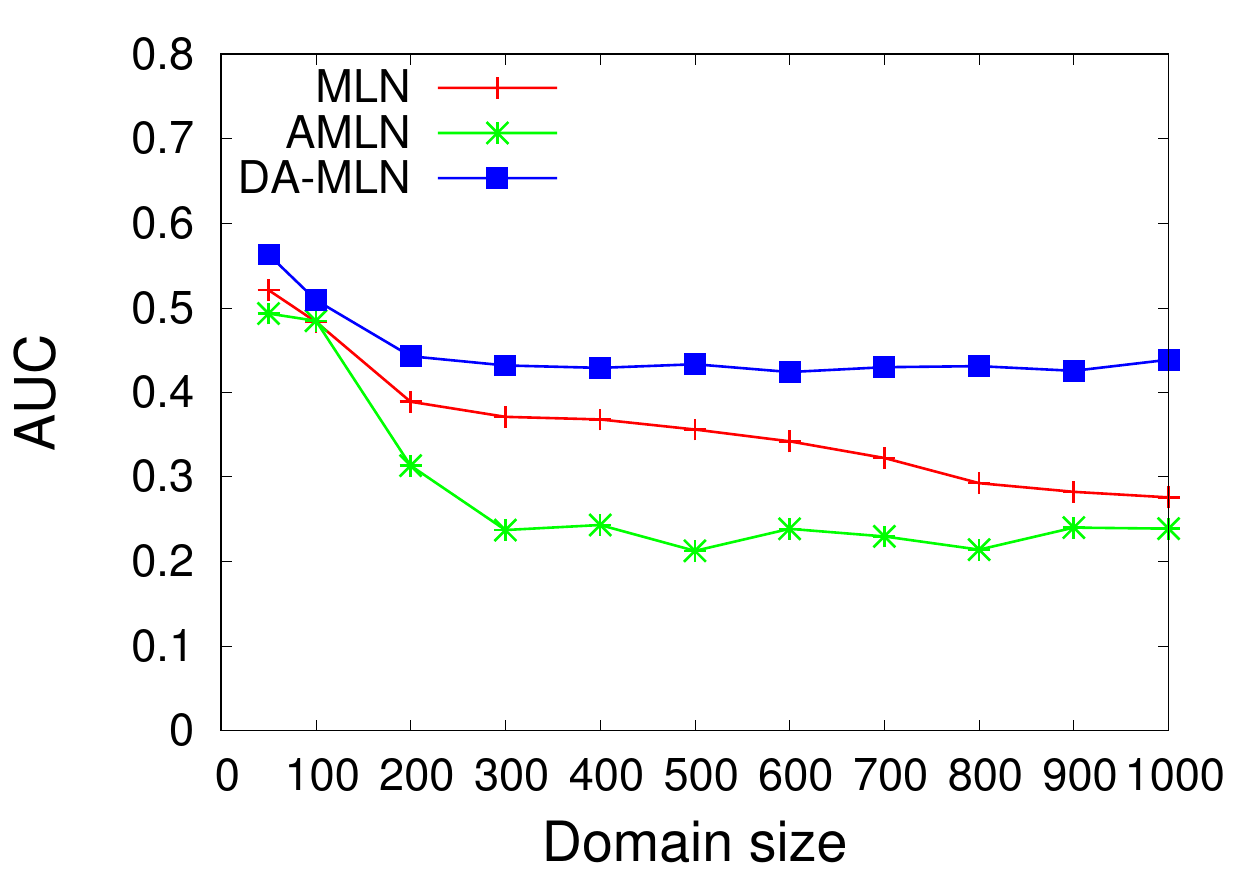}
 		\caption{Domain Size vs AUC (FS)} 
 		\label{fig:fs-auc}
 	\end{subfigure}
 	\caption{Results on FS dataset}
 \end{figure}

We used a set of randomly generated datasets with domain sizes $20,40,60,80,100$ for learning. For inference, we used randomly generated datasets of sizes varying from $50$ to $1000$. All the groundings of the \emph{Friends} predicate and randomly chosen $50\%$ groundings of \emph{Smokes} were set as evidence during inference. Figure~\ref{fig:fs-auc} plots the Area under the Precision-Recall curve (AUC) as we vary the test data sizes. While for smaller domain sizes all the algorithms perform equally well, performance
of AMLNs and MLNs drops drastically as the domain size increases. In contrast, DA-MLNs see much less drop in performance with increasing domain size, and perform significantly better than the competitors on larger domains. Test set log-likelihood shows a similar trend (omitted due to lack of space). We have also done some experiments on another real world dataset, and results are quite promising. We plan to publish these results in an extended conference version of the paper.

%% file: ack.tex
\vspace{-0.1 in}
\section{Acknowledgements}
Happy Mittal is being supported by the TCS Research Scholars Program.
Vibhav Gogate and Parag Singla are being supported by the DARPA Explainable Artificial Intelligence (XAI) Program with number N66001-17-2-4032. Parag Singla is being supported by IBM Shared University Research Award and the Visvesvaraya Young Faculty Research Fellowship by the Govt. of India. Vibhav Gogate is being supported by the National Science Foundation grants IIS-1652835 and IIS-1528037.

%Happy Mittal is being supported by the TCS Research Scholars Program. Parag Singla is being supported by a DARPA grant funded under the Explainable AI (XAI) program and also by the Visvesvaraya Young Faculty Fellowships by Govt. of India. Any opinions, findings, conclusions or recommendations expressed in this paper are those of the authors and do not necessarily reflect the views or official policies, either expressed or implied, of the funding agencies.